# Comparative Safety Performance of Autonomous- and Human Drivers: A Real-World Case Study of the Waymo One Service

Luigi Di Lillo*αβ, Tilia Godeφ, Xilin Zhou*, Margherita Atzei*, Ruoshu Chenφ, Trent Victorφ

*Swiss Reinsurance Company, Ltd, Reinsurance Solutions, α Research Affiliate with the Autonomous Systems Laboratory at Stanford University, β Research Affiliate with Frazzoli Group at ETH Zürich, φ Waymo LLC

**This study compares the safety of autonomous- and human drivers. It finds that the Waymo One autonomous service is significantly safer towards other road users than human drivers are, as measured via collision causation. The result is determined by comparing Waymo's third party liability insurance claims data with mileage- and zip-code-calibrated Swiss Re (human driver) private passenger vehicle baselines. A liability claim is a request for compensation when someone is responsible for damage to property or injury to another person, typically following a collision. Liability claims reporting and their development[1] is designed using insurance industry best practices to assess crash causation contribution and predict future crash contributions. In over 3.8 million miles driven without a human being behind the steering wheel in rider-only (RO) mode, the Waymo Driver incurred zero bodily injury claims in comparison with the human driver baseline of 1.11 claims per million miles (cpmm). The Waymo Driver also significantly reduced property damage claims to 0.78 cpmm in comparison with the human driver baseline of 3.26 cpmm. Similarly, in a more statistically robust dataset of over 35 million miles during autonomous testing operations (TO), the Waymo Driver, together with a human autonomous specialist behind the steering wheel monitoring the automation, also significantly reduced both bodily injury and property damage cpmm compared to the human driver baselines.**

This study examines whether the Waymo One™ ride-hailing service in San Francisco, CA and Phoenix, AZ exhibits better safety performance towards other road users than human drivers, as measured by real-world *auto third-party liability insurance claims data*, henceforth referred to as liability claims data. The Waymo Driver, the core of the Waymo One service, is a level 4 Automated Driving System (ADS) as defined in SAE J3016 (SAE 2021) and does not require a human driver behind the wheel when in Rider-Only (RO), "driverless" operation[2].

Valid "apples-to-apples" comparisons must overcome differences in collision reporting standards between autonomous and human driven vehicles, correct the underreporting in police-report data, use operational-design-domain-specific human driver comparison data, apply a statistical method to measure uncertainty, and should account for crash causation contribution (Blincoe et al., 2023; Victor et al., 2023; Blanco et al., 2016; Lindman et al., 2017; Bargman et al., submitted). Improperly controlled variations across collision datasets can lead to inflated or deflated collision statistics, and thus incorrect interpretations regarding safety.

Liability insurance claims offer a more comprehensive assessment than collision databases from police reports because (a) claims data have more consistent standards for reporting (Isaksson-Hellman et al., 2018), (b) claims data are demonstrated to have higher reporting frequency of safety-relevant crashes (Blincoe et al., 2023; Isaksson-Hellman & Lindman 2018), and (c) police reports do not capture non-collision related injuries (Mills et al., 2011) and capture fewer instances of injury claims (Isaksson-Hellman et al., 2018), see Figure 1.

---

[1] Claim count development is considered by reviewing known events. Future development (additional claim counts) is still possible from unreported and/or underreported claims. For Waymo, claims emergence months after the collision date is less likely due to the ability to detect event occurrence in a timely manner compared to human driven vehicles.

[2] cpuc-av-program-applications-guidance-20211026.pdf The CPUC application process makes the distinction between: The "Drivered AV Passenger Service" program allows for the provision of passenger service in test AVs with a driver in the vehicle; The "Driverless AV Passenger Service" program allows for the provision of passenger service in test AVs without a driver in the vehicle

Further, liability claims data uniquely capture information with regard to crash or injury causation contribution[3], since collision responsibility is directly determined during the liability claims adjudication process under the insurance industry best practices (Braver et al., 2004).

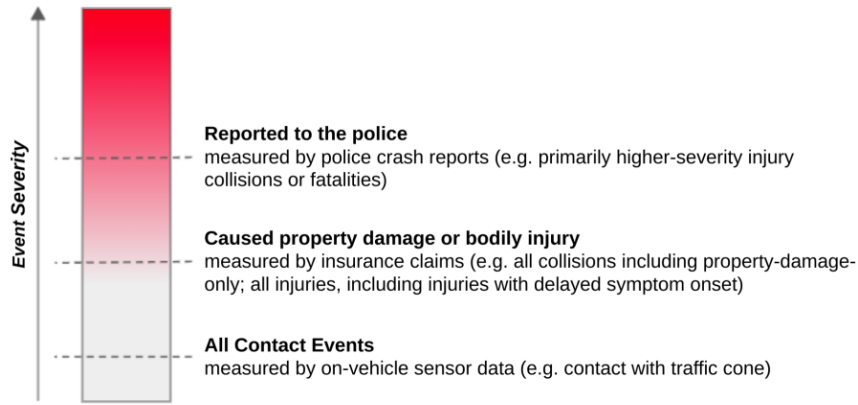

**Figure 1. Comparison of insights carried by different datasets.** The "All Contact Events" set contains all events, including those which are not safety-relevant. The "Reported to the police" set, on the other hand, is biased towards high severity events and lacks standardization in reporting.

In this study, we analyze claims filed under third-party liability (see footnote 3) insurance policies, which drivers are required to carry by law in California and Arizona[4], split by Property Damage Liability and Bodily Injury Liability coverages. Figure 2 shows the results of the comparison of liability insurance claims for bodily injury (left) and property damage (right) of Swiss Re human driver baselines and Waymo data. Note that Waymo's liability claims data used in this study is evaluated as of August 1, 2023.

---

[3] After a collision, the percentage of responsibility assigned to each party is determined by an insurance claims adjuster. Liability in the insurance context is distinct from the legal concept of fault. For the purposes of this study, we conservatively assume that the policyholder contributed to the collision giving rise to injuries or third party property damage if the claim was resolved with a liability payment by means of the insurance claims adjustment process. Also for the purposes of this study, we are using claims that are resolved, or likely to resolve with a liability payment, as a proxy for partial or full responsibility.

[4] Auto Insurance Requirements - California DMV

**Results**
- When Waymo vehicles were driven fully **manually**[5] by trained human drivers for 14,436,298 miles for data collection, bodily injury claims frequency was reduced by **45%** compared to the Swiss Re human driver baseline (0.55 vs 1.01 claims per million miles), but an overlap in 95% confidence intervals (*Manual$_{BI}$ 95% CI* [0.24, 1.09], *Baseline 95% CI* [1.00, 1.02]) indicates that there is insignificant or inconclusive evidence of this reduction. Property damage claims frequency was significantly reduced by **34%** (2.22 vs 3.34 claims per million miles). This result was confirmed by non-overlapping confidence intervals (*Manual$_{PDL}$ 95% CI* [1.52, 3.13], *Baseline 95% CI* [3.33, 3.36]).
- While driving 35,228,320 miles in **testing operations (TO)**[6] mode, the Waymo Driver, together with autonomous specialists[7], significantly reduced bodily injury claims frequency by **92%** (0.09 vs 1.09 claims per million miles), *TO$_{BI}$ 95% CI* [0.02, 0.25], *Baseline 95% CI* [1.08, 1.09]. Property damage claims frequency was significantly reduced by **95%** (0.17 vs 3.17 claims per million miles), *TO$_{PD}$ 95% CI* [0.06, 0.37], *Baseline 95% CI* [3.16, 3.18].
- While driving without a human behind the steering wheel in **RO** mode for 3,868,506 miles, the Waymo Driver reduced bodily injury claims frequency by **100%**, or zero claims, (0.00 vs 1.11 claims per million miles). The difference is statistically significant, indicated by the non-overlapping confidence intervals (*RO$_{BI}$ 95% CI* [0.000, 0.95], *Baseline 95% CI* [1.10, 1.12]). This provides strong evidence about the ADS' ability to reduce bodily injuries on public roads. Property damage claims frequency was significantly reduced by **76%** (0.78 vs 3.26 claims per million miles), as indicated by non-overlapping 95% CIs (*RO$_{PD}$ 95% CI* [0.16, 2.27], *Baseline 95% CI* [3.24, 3.27]).
- When **TO and RO** datasets were *combined*, totaling 39,096,826 miles, there was a significant reduction in bodily injury claims frequency by **93%** (0.08 vs 1.09 claims per million miles), *TO+RO$_{BI}$ 95% CI* [0.02, 0.22], *Baseline 95% CI* [1.08, 1.09]. Property damage claims frequency was significantly reduced by **93%** (0.23 vs 3.17 claims per million miles), *TO+RO$_{PDL}$ 95% CI* [0.11, 0.44], *Baseline 95% CI* [3.16, 3.18].

---

[5] In "Manual" mode, the Waymo vehicle is driven completely manually (i.e., without the ADS engaged) by an autonomous specialist (human driver) who is capable of reacting to a dynamic environment and operating vehicles under strict safety guidelines. For sake of clarity, a collision that follows after the autonomous specialist takes over control of the driving task does not count as a Manual claim, but as a TO claim, if the ADS performed a driving maneuver that placed the vehicle in the situation that led to the collision.

[6] In "TO" mode, the ADS is engaged to operate the vehicle under monitoring of trained human autonomous specialists. A collision is categorized under TO as long as the ADS was engaged at any time during the five seconds leading up to the impact. A collision could still be categorized under TO under the 5-second rule even if a human maneuver may have led to the collision, leading to a conservative estimate.

[7] Waymo One is assessed as a service which includes both the Waymo Driver safety performance as well as the autonomous specialist safety performance while in TO phase, as the claims outcomes are a product of both.

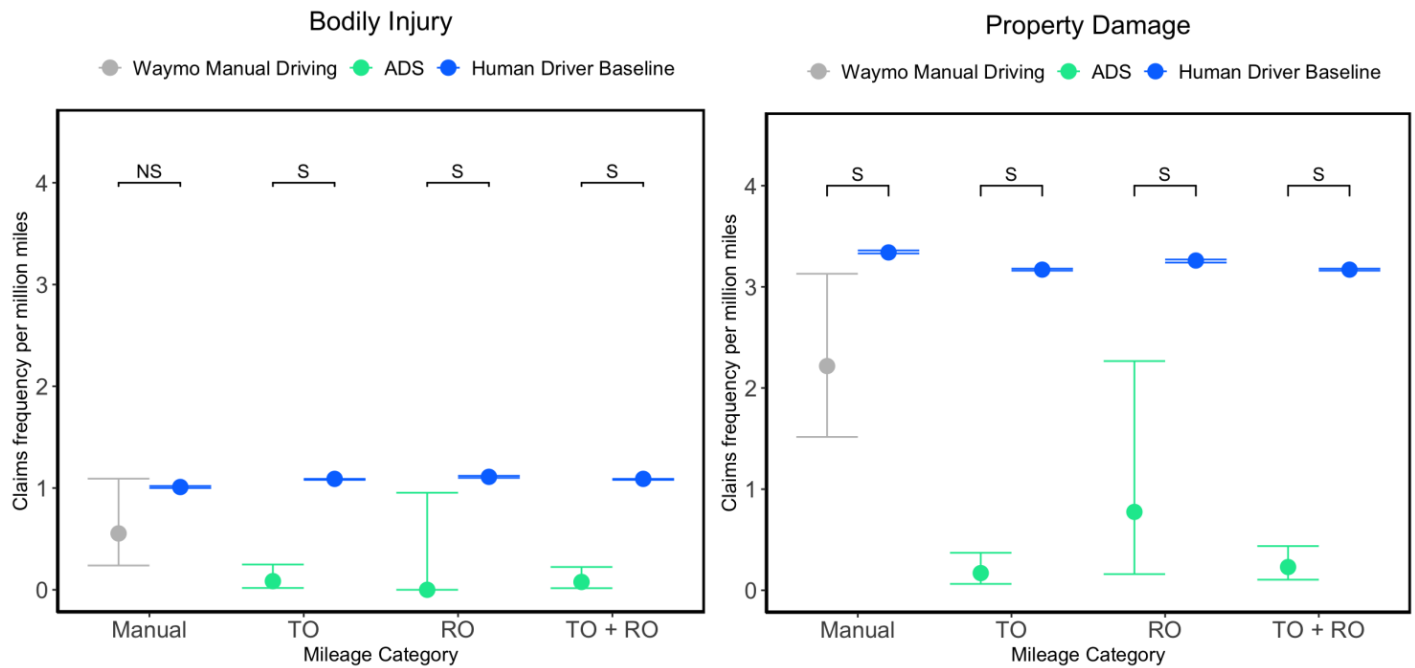

**Figure 2. Comparison of Swiss Re human driver baselines with Waymo liability insurance claims for bodily injury (left) and property damage (right).** S=significant (non-overlapping 95% CIs), NS=non-significant or inconclusive results (overlapping CIs). ADS=Waymo One Automated Driving System, Baseline=Swiss Re private passenger vehicle (human driver) baselines, calibrated for each mileage category. "Manual" is a mode in which the Waymo vehicle is driven manually (i.e., without the ADS engaged). Testing operations ("TO") is a phase of public road testing in which the ADS is engaged under monitoring of a trained autonomous specialist (human driver) who is seated in the driver's seat and can take over the driving task at any time. TO collisions were included even if they occurred up to 5 seconds after the autonomous specialist took over control (disengaged the ADS). Rider-only ("RO") is a mode where the ADS operates the vehicle without any human behind the steering wheel. Testing operation and Rider-only ("TO+RO") is the combination of the TO and RO datasets.

**Methods**

The human driver baselines are based on Swiss Re's property damage liability (PD) and bodily injury liability (BI) claims data from 2016 to 2021, from over 600,000 claims and over 125 billion miles of exposure. Property Damage Liability insures against damages that at-fault drivers cause to other people's vehicles and property in crashes. Bodily Injury Liability insures against medical, hospital, and other expenses for injuries that at-fault drivers inflict on occupants of involved vehicles or others on the road. Since insurance claims data includes information on cost severity rather than injury severity, we do not differentiate levels of injury severity. In addition, as one collision can lead to both bodily injury and property damage claims, such collisions are included in both BI and PD comparisons.

The baseline was calibrated using both mileage (driving exposure) and zip-code (geographic region). For the mileage- and zip-code-calibrated baselines, only the claims associated with vehicles registered to addresses (i.e., where the insured resides) within Waymo's operating zip codes in San Francisco and the Phoenix metropolitan region were included. When doing traditional territorial ratemaking, the best proxy estimate of the claim frequency in an area is obtained by observing the frequency for residents of that area, given that the majority of claims happen within a small radius of residency. Thus, this zip-code mismatch (i.e., zip code of the collision vs. zip code of registered vehicle address) is expected to have an immaterial impact on the frequency estimate.

Since Waymo's claims exposure is measured using mileage, in order to produce a valid human baseline for comparison, we convert the number of exposure *years* contained in the human driven dataset to the number of exposure *miles*. To do so, we estimate the annual vehicle miles traveled (VMT) for one vehicle[8]. Annual VMT per vehicle is estimated at a yearly and regional (state or city) granularity.

Our estimation method uses aggregate mileage data to then calculate a per-vehicle average. For the mileage data, we use VMT statistics provided by the Federal Highway Administration (FHWA) which reports the total monthly VMT by state[9] and total annual VMT by urbanized area[10]. These statistics are based on individual state reports of traffic data counts collected through permanent automatic traffic recorders on public roadways. In addition, for per-vehicle statistics, we use FHWA's annual vehicle registration statistics[11] and US Census Data[12]. For both San Francisco and Phoenix and for each of the six coverage years included in the baseline (2016-2021), we produce two VMT per vehicle estimates: one estimate which draws from state VMT data, and another estimate based on VMT data per urbanized area. For the analysis presented above, we chose to use estimates based on state VMT data because it yielded a lower (more conservative) baseline frequency, given that the state average annual miles per vehicle are higher than those of the urbanized area.

For the mileage- and zip-code-calibrated baselines, claim frequencies are independently calculated for vehicles within the San Francisco and the Phoenix metropolitan area. To build the baselines for comparison, these frequencies are then mixed according to Waymo's mileage distribution across Phoenix and San Francisco. The Waymo data included in this study is from miles driven and claims occurred and reported in the period from 1/1/2018 to 8/1/2023. Since the different Waymo mileage categories (RO, TO, TO + RO, and Manual) have different Phoenix-San Francisco mileage distributions, separate Human Driven Vehicle (HDV) baselines are calculated for each of the four mileage categories.

---

[8] A generally accepted estimate for annual VMT per vehicle in the US is approximately 12,000 miles. However, due to variations in driving patterns across different US cities and states, we separately estimate annual VMT per vehicle for each region (city or state) within the baseline.

[9] Office of Highway Policy Information - Policy | Federal Highway Administration (dot.gov)

[10] Table HM-71 - Highway Statistics 2020 - Policy | Federal Highway Administration (dot.gov) An urbanized area is defined as an area with 50,000 persons that at a minimum encompasses the land area delineated as the urbanized area by the U.S. Census Bureau.

[11] Table MV-1 - Highway Statistics 2020 - Policy | Federal Highway Administration (dot.gov)

[12] Data (census.gov)

For significance testing, we conclude that there is a significant difference between claims frequencies if the 95% confidence intervals of Waymo's operations and their corresponding human baselines do not overlap. Namely, that there is a difference even when accounting for statistical uncertainty due to the small mileage volume. For the human baseline, due to the large sample, we use a normal approximation confidence interval and take the mileage distribution between San Francisco and Phoenix into account when computing the standard error. For Waymo's operations, due to the smaller sample and exposure size, confidence intervals are calculated using the Poisson Exact Method (Garwood, 1936)[13].

**Discussion**

In this study, the baseline for comparison was derived from a human population of insured drivers that reside in the same zip codes as the Waymo's service. Benefits of this population include its size and robustness, which lends itself to narrower confidence intervals. In addition, the selected baseline population is likely the population that may use Waymo services instead of driving themselves.

A limitation of the selected human baseline is that the location of crashes that generate claims is not known, which limits the ability to filter claims based on Waymo's Operational Design Domain (ODD). As a result, whereas the Waymo ODD largely does not include freeway driving, the human database includes miles driven and claims which occur on freeways. Due to variations in collision frequency per million miles between freeways and non-freeways, this may have led to a baseline which may be more conservative than a roadway-matched baseline. Due to the fact that in territorial ratemaking the frequency observed for residents in a specific area is considered to be the best proxy estimate of the claim frequency in that area, the impact of these differences is expected to be negligible. For future studies, we plan to investigate other populations and methods to subset the data to generate comparisons.

**Closing remarks**

This study introduces the use of private passenger vehicle (human driver) liability insurance claims data to construct performance baselines to benchmark the safety performance of autonomous vehicles. We demonstrate that the Waymo One service is substantially safer (i.e., regarding % reduction in number of liability claims), compared to robust and highly significant private passenger vehicle baselines established by Swiss Re, with over 600,000 claims and over 125 billion miles of exposure, and calibrated to match Waymo's mileage distribution across operating locations.

This method overcomes existing challenges facing autonomous and human driver crash rate comparisons and can be applied within the industry to assess the safety performance of additional autonomous vehicle deployments.

---

[13] Note that the baseline's confidence intervals are created using estimated annual mileages, and different estimation assumptions will yield different point estimates and confidence intervals.

**Contributors**: Verena Brufatto, Solutions Data Analyst, and Federica Capparelli, Data Scientist, have been instrumental in creating the baselines and analytics around this work. Binbin Li, Risk Analyst, has contributed greatly to putting together the statistical analysis and ensuring data quality. Ross Amend, Claims Expert, was instrumental in ensuring data accuracy, Matt Glascock, Motor Pricing Expert, has offered invaluable support by sharing his expertise on motor claims data and by reviewing the methodological assumptions and the analytical work.

**Acknowledgements**: Andrea Biancheri, Senior Pricing Actuary, and Cristiano Misani, Lead Advanced Scoring, have been critical in challenging the analysis and the results.